\documentclass{article}

\usepackage{arxiv}

\usepackage[utf8]{inputenc} 
\usepackage[T1]{fontenc}    
\usepackage{hyperref}       
\usepackage{url}            
\usepackage{booktabs}       
\usepackage{amsfonts}       
\usepackage{nicefrac}       
\usepackage{microtype}      
\usepackage{url}

\usepackage{lipsum}         
\usepackage{graphicx}
\usepackage{natbib}
\usepackage{doi}
\usepackage{color}
\usepackage{amsbsy}

\usepackage{graphicx}
\usepackage{epstopdf}

\usepackage{algorithm}
\usepackage{algorithmicx}
\usepackage{algpseudocode}
\usepackage{amsmath}
\usepackage{array}

\usepackage{cleveref}       

\floatname{algorithm}{Algorithm}

\title{High-Quality Real Time Facial Capture Based on Single Camera}

\author{ {Hongwei Xu}\thanks{Corresponding author} \\
	FACEGOOD \\
	\And
	{Leijia Dai} \\
	FACEGOOD \\
	\And
	{Jianxing Fu} \\
	FACEGOOD \\
	\And
    {Xiangyuan Wang} \\
    FACEGOOD \\
    \And
	{Quanwei Wang} \\
	FACEGOOD \\
}

\renewcommand{\undertitle}{}

\hypersetup{
pdfauthor={Hongwei Xu},
pdfkeywords={Facial Capture, Machine Learning, Knowledge Distill},
}

\begin{document}
\maketitle

\begin{abstract}
	We propose a real time deep learning framework for video-based facial expression capture. Our process uses a high-end facial capture pipeline based on FACEGOOD\footnote{https://www.facegood.cc/} to capture facial expression. We train a convolutional neural network to produce high-quality continuous blendshape weight output from video training. Since this facial capture is fully automated, our system can drastically reduce the amount of labor involved in the development of modern narrative-driven video games or films involving realistic digital doubles of actors and potentially hours of animated dialogue per character. We demonstrate compelling animation inference in challenging areas such as eyes and lips.
\end{abstract}


\section{Introduction}
One of the hottest areas of computer vision and graphics is capturing facial expression in real time and drive a virtual digital human to act like a specific actor. This technology enables any user to control the facial expression of a digital avatar in real time. James Cameron's sci-fi blockbuster "Avatar" remains in first place with nearly 3 billion dollars. Emotion-capture technology has captured the facial details of Thanos's cast in The Avengers to the delight of millions of viewers. Avatar and Thanos used facial expression capture technology, which is fascinating.

Facial expression capture technology is mainly divided into landmark driven, point cloud driven, sound driven and image based. These types of methods will then be described in more detail.

Humans express their inner feelings through facial expression, such as happiness or anger. How to make the computer automatically recognize expressions is an important research field, which has attracted many scholars to devote their energy and time to him. Essa et al.\citep{essa1996modeling} show that facial animation methods are also designed for basic video input, but their heavy dependence on optical flow or feature tracking may lead to instability. Facial expressions often play an indispensable role in animation, games, films and television production.

In the movies, the creature animation of characters with strange facial structures, such as King Kong and Gollum, are generated through this technology. How to use the expression data of a single actor to drive different face models has attracted great attention in the field of face simulation. In the field of motion capture, the motion data of different facial expressions are mapped and registered. The feature reconstruction method depends on the training data under different occlusion conditions, but occlusion positions and types are too many to control. Li et al.\citep{li2018patch} designed patch gated revolution neutral network(pg-cnn) for sensing occlusion, which can automatically perceive the occlusion area of the human face, focus on the divided 24 non-occlusion sub-regions with more feature information, and input the 24 sub-areas into an attention network to obtain weighted cascaded local features. Li et al.\citep{li2018occlusion} further extended the pg-cnn method by introducing global candidate units to supplement the global information of face images for expression recognition. However, these methods of selecting sub-regions based on face key points are not accurate in face images with occlusion.

Recent studies use different network structures and preprocessing methods. In the field of expression recognition, the rapid development of deep learning has prompted researchers to use deep neural network to develop facial expression recognition In recent years, researchers have proposed various novel expression recognition models and algorithms to improve the efficiency of expression recognition and reduce the error of expression recognition in special scenes. So they have shown that convolutional neural network can also extract features and classify facial expression recognition. Lewis et al.\citep{lewis2014practice} show that, despite the simplicity of the blendshape approach, there remain open problems.

There are many models based on point cloud. Iordanis et al.\citep{mpiperis2008bilinear} introduces a novel model-based framework for establishing correspondence among 3d point clouds of different faces. Xiao et al.\citep{xiao2004real} study the representational power of Active Appearance Models which are popular generative models and show that they can model anything a 3d Morphable Model can, but possibly require more shape parameters. In 2009, Weise et al.\citep{weise2009face} produces A real-time structured light scanner which is low-cost can provide dense 3d data and texture. With the emergence of RGBD camera, depth data can be obtained by researchers. The physical distance is obtained by dividing the pixel value of the depth map obtained by the RGBD camera by the scale map.

Hao et al.\citep{li2013realtime} presented a calibration-free facial performance capture framework based on a sensor with video and depth input in 2013. Samuli et al.\citep{laine2017production} produces high-quality output based on a convolutional network, including self-occluded regions, from a monocular video. Thibaut et al.\citep{weise2009face} propose a system for live puppetry that allows transferring an actor’s facial expression onto a digital 3d character in 2009. In 2011, Thibaut et al.\citep{weise2011realtime} introduce a novel face tracking algorithm that combines geometry and texture registration. And it effectively maps low-quality 2d images and 3d depth maps to realistic facial expression. Martin et al.\citep{klaudiny2017real} achieve markerless facial performance capture from multi-view helmet camera data, employing an actor-specific regressor in 2017. Derek et al.\citep{bradley2010high} introduced a purely passive facial capture approach that uses only an array of video cameras, but requires no template facial geometry, no special makeup or markers, and no active lighting.

Audio-driven facial simulation technology is also attracting much research. Because little has been done to model expressive visual behavior during speech, Yong et al.\citep{cao2005expressive} address this issue using a machine learning approach that relies on a database of speech-related high-fidelity facial motions. Tero et al.\citep{karras2017audio} drive real-time 3D facial animation by audio input with low latency. Neural network based end-to-end models suffer from slow inference speed, and the synthesized speech is usually not robust, i.e., some words are skipped or repeated. But, Yi et al.\citep{ren2019fastspeech} extract attention alignments from an encoder-decoder based teacher model for phoneme duration prediction

Image-based systems are completely different from model-based system, such as the one introduced by Iordanis as mentioned above. Kang et al.\citep{liu2008robust} presented an image-based facial animation system using Active Appearance Models  for precisely detecting feature points in the human face. Kang et al.\citep{liu2009optimization} concatenates appropriate mouth images from the database such that they match the spoken words of the talking head in 2009. In 2011, they\citep{liu2011realistic} presents an image-based talking head system that is able to synthesize realistic facial expression accompanying speech, given arbitrary text input and control tags of facial expression. Sarah et al.\citep{taylor2016audio} present a sliding window deep neural network that learns a mapping from a window of acoustic features to a window of visual features. Chen et al.\citep{chen2013accurate} proposed an image-based 3d nonrigid registration process. In the model presented by Cao et al.\citep{cao20133d}, the 3D positions of facial landmark points are inferred by a regressor from 2d video frames of a web camera. They capture 150 individuals aged 7–80 from different backgrounds by RGBD camera\citep{cao2013facewarehouse}. Chen et al.\citep{cao2014displaced} proposed an approach that does not need any calibration for each individual user. But it demonstrates a level of robustness and accuracy on par with state of the art techniques that require a time-consuming calibration step. Chuang et al.\citep{chuang2002performance} use a combination of motion capture data and blendshape interpolation to create facial animation. In the process of traditional expression animation, the realistic effect of animation often needs large manual intervention. Therefore, authenticity and efficiency are still the main indexes of current facial expression simulation research. Motion capture technology is a new data acquisition means developed in recent years, which can record and restore the performer's motion in real time. Face motion capture technology applies motion capture technology to the production of facial expression animation. Therefore, in recent years, motion capture technology is in the research field of facial expression simulation Domain has attracted more and more attention. Assia et al.\citep{khanam2007intelligent} presents a new approach to add flexibility and intelligence to Performance Driven Facial Animation by means of context-sensitive facial expression blending. Salil’s \citep{deena2009speech} facial animation is done by modeling the mapping between facial motion and speech using the shared Gaussian process latent variable model. In the research of Meng et al.\citep{meng2019lstm}, the embeddings are fed into an Long Short Term Memory Network (LSTM) network to learn the deformation between frames.

Humans express their inner feelings through facial expression, such as happiness or anger. How to make the computer automatically recognize expression is a hot research field, which has attracted many scholars to devote their energy and time to him. Extracting image features with strong robustness and representation ability is the key of expression recognition system. Various 3d models for human faces have been used in computer graphics and computer vision. The model introduced by Marc et al.\citep{habermann2019livecap} is the first real-time monocular approach for full-body performance capture. Kettern et al.\citep{kettern2015temporally} proposed a model that supports any number of arbitrarily placed video cameras.

Facial expression simulation has attracted much research effort in the fields of virtual reality, cognitive science, human-computer interaction interface design and online video conference. Authenticity is one of the key evaluation standard of expression simulation. Face animation methods are also designed for basic video input, but their heavy dependence on optical flow or feature tracking may lead to instability. Facial expression often play an indispensable role in animation, games and film and television production. Zhang et al. \citep{zhang2014random} extracted a set of Gabor based face templates by Monte Carlo method in 2014 and transformed them into template matching distance features. The distance feature of template matching depends on the template selection in a specific expression dataset, so it is difficult to guarantee the performance across datasets and does not have good generalization ability. Another feature reconstruction method is to learn a generation model, which can reconstruct a complete face from the occluded face\citep{pan2019occluded}.

Disadvantages of RGBD-based systems as following: computing-time consuming, expensive equipment, and no eyeball performance. Similarly, all kinds of audio-driven method cannot drive the eyeball correctly. But our distill facial capture network(DFCN) involves the performance of the eyes. A lot of them are model-based, whereas we are image-based. Further, knowledge distill is introduced into the regression algorithm to improve the expression ability of the network.

Salient contributions in this paper can be summarized by

1. A framework for real-time facial capture from video sequences to blendshape weight and 2d facial landmark is established.

2. An adaptive regression distillation(ARD) framework is proposed, which can filter mislabeled data and ensure that student network is trained on the right track.

The rest of this paper is organized as follows. In Section \ref{sec:Problem Formulation}, the facial capture problem is mathematically formulated. The scheme is developed in Section \ref{sec:End-to-end Algorithm}. In Section \ref{sec:Experimental Results}, experimental results and discussions are presented. Conclusions are drawn in Section \ref{sec:Conclusion}.

\begin{figure}
\centering
  \includegraphics[scale=0.8]{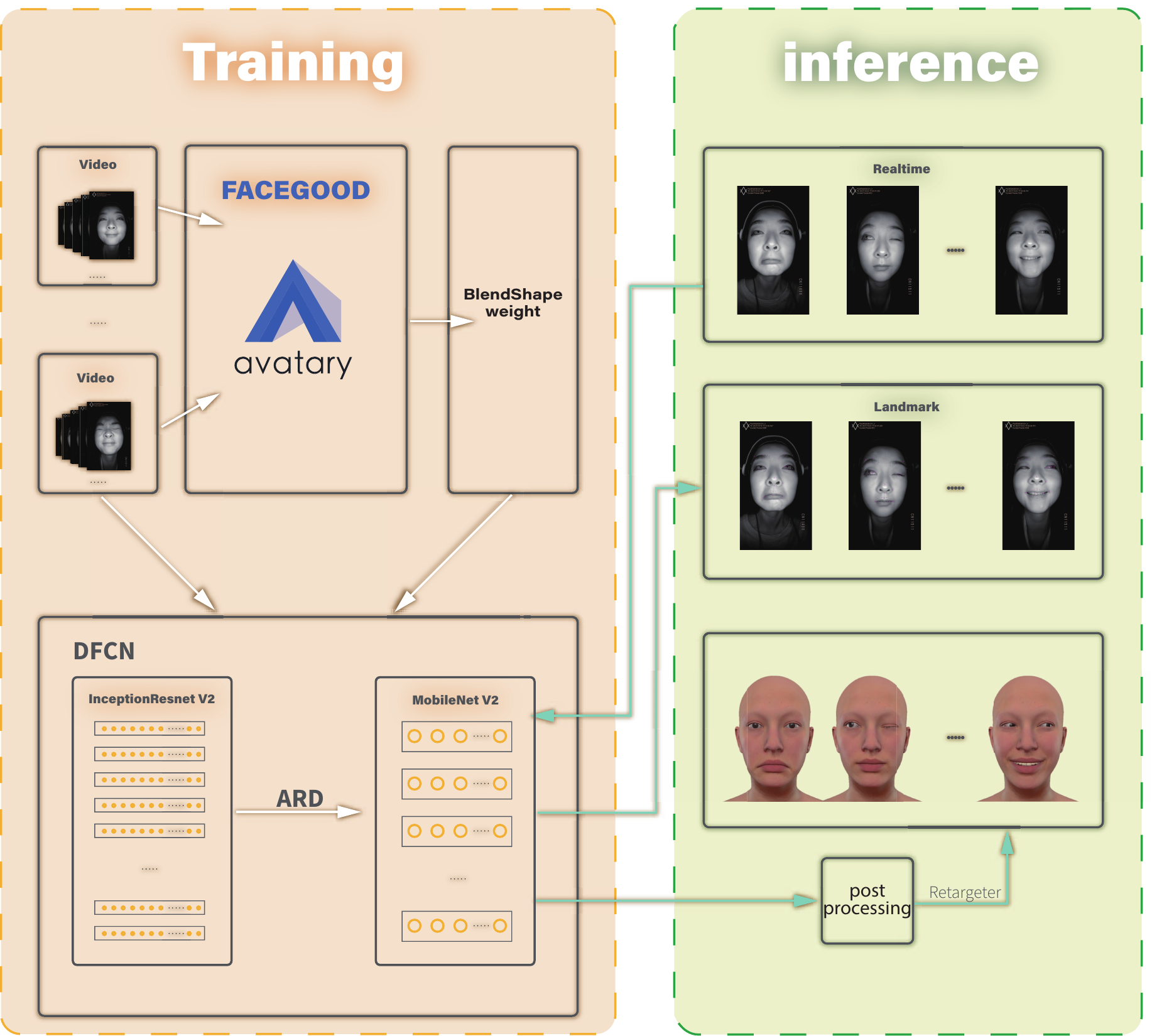}\\
  \caption{Our deep learning-based facial performance capture framework is divided into a training and inference stage. The goal of our system is to reduce the amount of footage that needs to be processed using labor-intensive production-level pipelines.}\label{main flow chart}
\end{figure}

\section{Problem Formulation}
\label{sec:Problem Formulation}

In general, as shown in Fig. 1, use an infrared camera which can avoid the influence of light on the picture to capture the image data of the face, and then use the end-to-end method to obtain the corresponding blendshape weight. In addition, in order to avoid discontinuities between frames, a hybrid smoothing method is adopted.

For 3d shapes that need to be driven, we use a Dynamic Expression Model(DEM) based on a set of blendshape meshes. Similar to [Weise et al. 2011], we represent the 3d facial mesh $ \pmb{F} $ as a linear combination of expression blendshapes $\pmb{B} = [\pmb{b}_0, ..., \pmb{b}_n]$:
\begin{equation}
\pmb{F} = \pmb{B}\pmb{e}^T
\label{equ:1}
\end{equation}
where $\pmb{e} = [e_0, ..., e_n]$ is the expression coefficients. As commonly assumed in blendshape models, $\pmb{b}_0$ is the neutral face, and nonneutral blend weights $e_i, 1 \leq i \leq n$ are bounded between 0 and 1. All blend weights must sum to 1, leading to $e_0 = 1 - \sum\nolimits_{i=1}^ne_i$ . For simplicity of description, we ignore the dependent $e_0$ in the following equations and discussions, and assume the correct value is computed on demand.

Our blendshape model is based on the ArKit\citep{arkit}, which contains 52 action units (i.e., $n = 52$) that mimic the combined activation effects of facial muscle groups. This blendshape model adequately describes the expression of the human face.

In the context, our algorithm can be expressed as the following:
\begin{equation}
DFCN(\pmb{I}) = (\pmb{e, S})
\label{equ:2}
\end{equation}
where the 2d facial shape $\pmb{S}$ is thus represented by the set of all 2d landmarks $s_k$, $\pmb{I}$ is input image.

\section{DFCN Algorithm}
\label{sec:End-to-end Algorithm}
In this section, by directly obtaining the weight of the corresponding blendshape and the 2d landmarks according to the ordinary image, DFCN algorithm is created, and it can resist the influence of different intensities of light and jitter from the outside world. Our pipeline is outlined in Fig. \ref{main flow chart}.

\subsection{Network Architecture}
As input for the network, we take the $1920\times1080$ video frame
from the camera, crop it with a fixed rectangle so that the face remains in the picture, and scale the remaining portion to $160\times160$ resolution. Furthermore, we normal the image, resulting in a total of 25600 scalars to be fed to the network. The resolution may seem low, but numerous tests confirmed that increasing it did not improve the results.

Our convolutional network is based on the InceptionResNetv2\citep{szegedy2017inception}, but cancels any activation functions in the final output. Although InceptionResNetv2 can extract image features very well, it cannot be calculated quickly in the CPU. Therefore, InceptionResNetv2 will be used as a teacher network to perform knowledge distillation on the student network, i.e., MobileNetv2, in order to achieve the role of model compression.

\subsection{Post-processing}
In terms of continuously ensuring the continuity of the output of the frame, a hybrid filtering method has been reached.

Kalman filter\citep{kalman1960new} is an optimal recursive data processing algorithm. The estimated value can be either stationary or non-stationary. Only the process noise, measurement noise and the statistical characteristics of the current system state need to be considered in the calculation. So the spatial complexity of the calculation is small.
The system equation is:
\begin{equation}
\pmb{x}_k = \pmb{Gx}_{k-1} + \pmb{Qu}_k + \pmb{w}_{k-1}
\label{equ:3}
\end{equation}
where $\pmb{x}$ is the state of the system. $\pmb{G}$ is the system matrix. $\pmb{Q}$ and $\pmb{u}$ is the control of state. $\pmb{w}$ is the process noise.
And the measurement equation is:
\begin{equation}
\pmb{z}_k = \pmb{Hx}_k + \pmb{v}_k
\label{equ:4}
\end{equation}
where $\pmb{z}$ is the thermometer reading and $\pmb{H}$ is the matrix of observations. In addition, $\pmb{v}$ is the measurement noise.

In Savitzky-Golay(SG) filter algorithm\citep{press1990savitzky}, the radius of sliding window and the order of polynomial are specified by the user. And the polynomial order must be less than the window radius. In addition, the window length must be odd.

There are $2n + 1$ expression like this: take time $t$ as an example:
\begin{equation}
x_t = a_0 + a_1  t + a_2  t^2 + ... + a_{k-1} t^{k-1}
\label{equ:5}
\end{equation}
where $k-1$ is the order of the polynomial used for fitting, $t$ is the time, and a is the parameter need to be figured out.
Therefore, $2n + 1$ of the above expression are written in the form of a matrix is:
\begin{equation}
\pmb{X}_{(2n+1) \times 1} = \pmb{H}_{(2n+1) \times k} + \pmb{A}_{k \times 1} + \pmb{E}_{(2n+1) \times 1}
\label{equ:6}
\end{equation}
where $\pmb{A}$ is the parameter vector need to be computed, and $\pmb{E}$ is the random noise column vector.
The subscript of the above formula refers to the dimention of each matrix.
Through the least square method, the following can be obtained:
\begin{equation}
\pmb{A} = (\pmb{H}^{T} \pmb{H})^{-1} \pmb{H}^{T} \pmb{X}
\label{equ:7}
\end{equation}
Then the SG filtering result is:
\begin{equation}
\pmb{P} = \pmb{H A} = \pmb{H} (\pmb{H}^{T} \pmb{H})^{-1} \pmb{H}^{T} \pmb{X}
\label{equ:8}
\end{equation}


\subsection{Training}

\subsubsection{Data Acquisition}
\begin{figure}
\centering
  \includegraphics[scale=1.0]{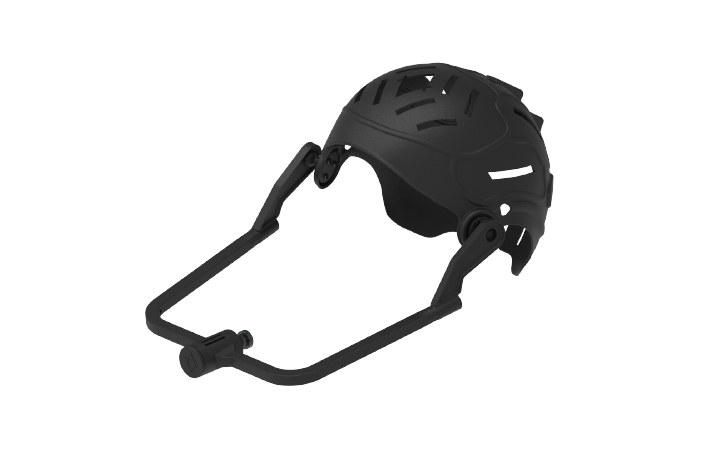}\\
  \caption{Facegood P1 Helmet.}\label{p1}
\end{figure}

In order to avoid the influence of light on the image, the FACEGOOD P1 equipment(Fig. \ref{p1}) is used for image acquisition. For each actor, the training set consists of four parts, totaling approximately 5-6 minutes of footage. The composition of the training set is as follows.

\emph{Standard Expression.} In order to capture the maximal extents of the facial motion, a single range-of-motion shot is taken where the actor goes through a pre-defined set of extreme expression. These include but are not limited to opening the mouth as wide as possible, moving the jaw sideways and front as far as possible, pursing the lips, and opening the eyes wide and forcing them shut. Farther, unlike the range-of-motion shot that contains exaggerated expression, this set contains regular FACSlike expression such as squinting of the eyes or an expression of disgust. These kind of expression must be included in the training set as otherwise the network would not be able to replicate them in production use.

\emph{Special Expression.} This set needs to increase the actors' all-out distorted expression, so that the network can better replicate these dramatic performances.

\emph{Speak Normally.} This set leverages the fact that an actor’s performance of a character is often heavily biased in terms of emotional and expressive range for various dramatic and narrative reasons. This material is composed of the preliminary version of the script, or it may be otherwise prepared for the training to ensure that the trained network produces output that stays in character.

\emph{Speak Exaggeratedly.} This set attempts to cover the set of possible facial motions during an exaggerated speech for a given target language.

Next step, the Avatary \footnote{https://www.avatary.cc/} software will be used to generate the blendshape weights from this videos.

\begin{figure}[t]
\centering
  \includegraphics[scale=0.65]{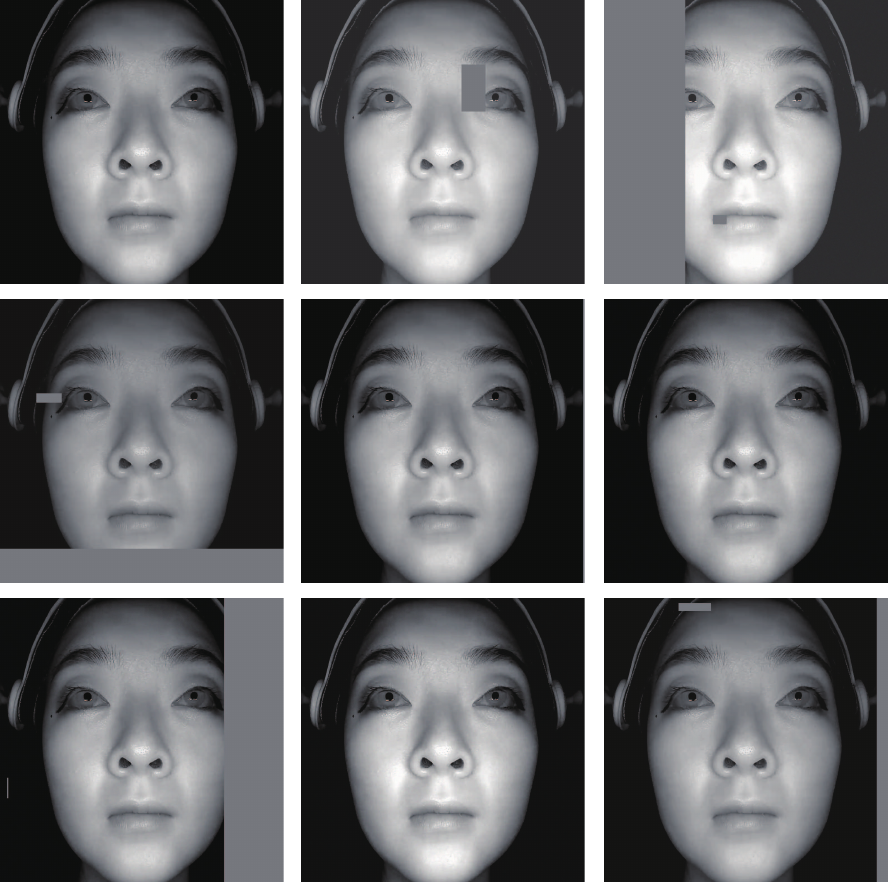}\\
  \caption{Examples of augmented inputs presented to the network during training.}\label{augment}
\end{figure}

\subsubsection{Data Augmentation}
We perform several transformations to the input images during training in order to make the network resistant to variations in input data. These transformations are executed on CPU concurrently with network evaluation and training that occurs on the GPU. Augmentation is not used when evaluating the validation loss or when processing unseen input data in production use. Examples of augmented input images are shown in Fig. \ref{augment}.

The main transformations are geometric transformations, we vary the brightness, random padding and contrast of the input images during training, in order to account for variations in lighting over the capture process.

\subsubsection{Training Parameters}
We train the teacher network for 400 epochs using the Adam\citep{kingma2014adam} optimization algorithm with parameters set to values recommended in the paper. The learning rate is ramped up using a geometric progression during the first training epoch, and then decreased according to $ 1 / \sqrt{t}$ schedule. During the last 60 epochs we ramp the learning rate down to zero using a smooth curve, and simultaneously ramp Adam $\beta_1$ parameter from 0.9 to 0.5. The ramp-up removes an occasional glitch where the network does not start learning at all, and the ramp-down ensures that the network converges to a local minimum. Minibatch size is set to 1024, and each epoch processes all training frames in randomized order.

\begin{equation}
q_i=\frac{e^{z_i/T}}{\sum_j{e^{z_i/T}}}
\label{equ:9}
\end{equation}
This equation suggests that the exponent of e in the sigmoid function is divided by the temperature $T$ to make the it smaller. The effect of this is to make the categories of sigmoid output smoother and let the teacher transfer the soft-target knowledge to the students, which can help students learn better.
Loss $L_{distll}$ is defined as:
\begin{equation}
L^{classification}_{distill} = \alpha T^2 K(\frac{\pmb{O}_{student}}{T}, \frac{\pmb{O}_{teacher}}{T}) + (1 - \alpha) M(\pmb{O}_{student}, \pmb{y})
\label{equ:10}
\end{equation}
where $K$ is the the Kullback-Leibler(KL) divergence operator, and $M$ is the CrossEntropy of student and correct label. And $\alpha$ is determined by the researcher.

\begin{figure}[t]
\centering
  \includegraphics[scale=0.65]{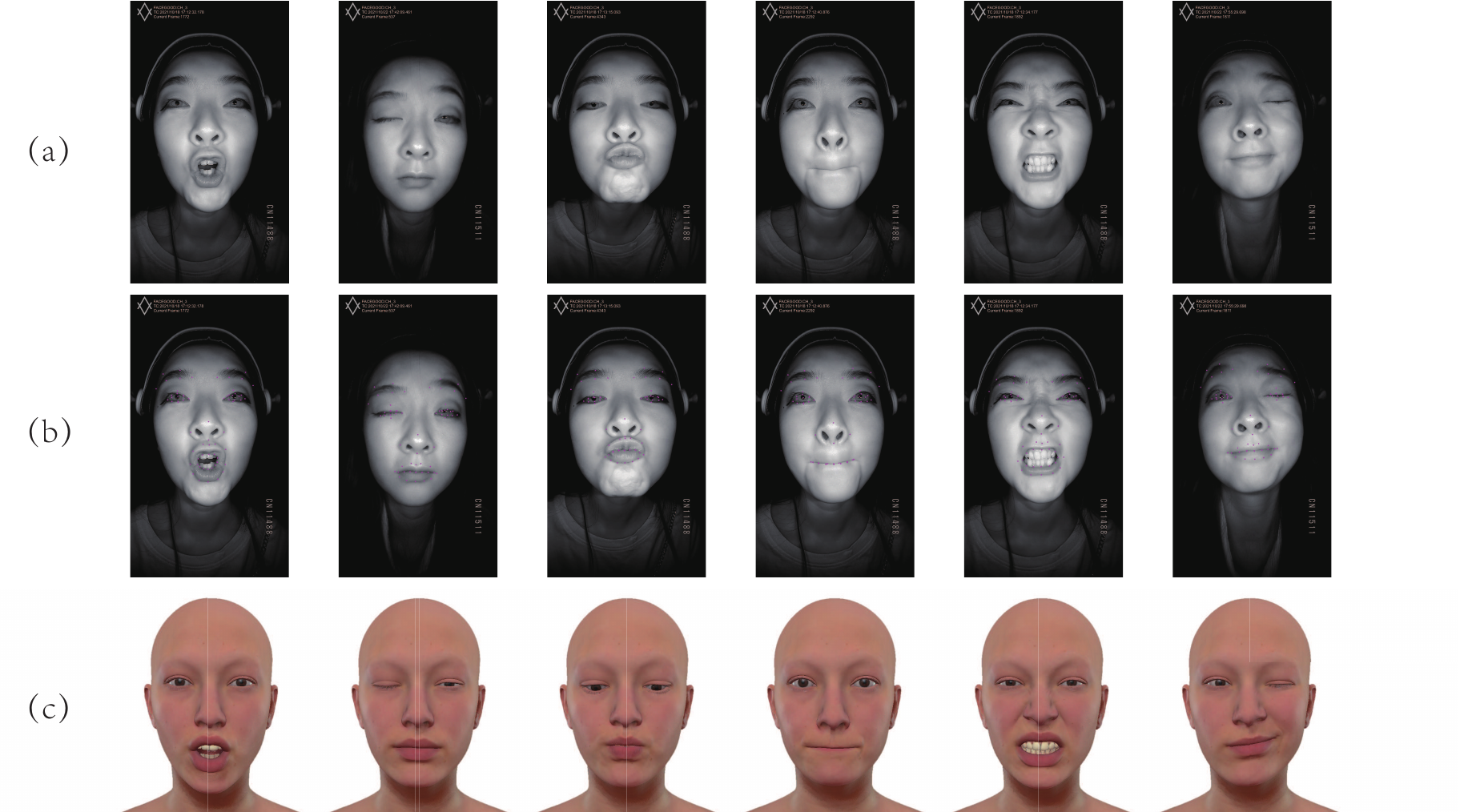}\\
  \caption{Three rows showing the raw expression, the expression with 2d landmarks, and the expression of metahuman driven by human.}\label{result}
\end{figure}

However, the current distillation is mainly used in classification tasks. By increasing the temperature, teachers can output the soft knowledge of "6 is not only like 6, 6 is also like 4", so as to enhance the generalization ability of students' models. There are only two papers on the application of distillation to regression tasks.
\begin{equation}
 L_{distill}=\left\{
\begin{array}{rcl}
||\pmb{O}_{student}-\pmb{O}_{teacher}||^2_2       &      & {if \ \ ||\pmb{y}-\pmb{O}_{teacher}||_2 > \mu}\\
f(\pmb{O}_{student}, \pmb{m})                     &      & {if \ \ ||\pmb{y}-\pmb{O}_{teacher}||_2 < \mu}\\
\end{array} \right.
\label{equ:11}
\end{equation}
where
\begin{equation}
f(\pmb{O}_{student}, y)=\left\{
\begin{array}{rcl}
v ||\pmb{O}_{student}, \pmb{m}||^2_2       &      & {if \ \  ||\pmb{O}_{student}-\pmb{m}||^2_2 + b > ||\pmb{O}_{teacher}-\pmb{m}||^2_2}\\
||\pmb{O}_{student}, \pmb{m}||^2_2      &      & otherwise\\
\end{array} \right.
\label{equ:12}
\end{equation}
Our idea is when the point is an outlier, then let the students learn from the teacher instead of the dirty data. On the contrary, if it is not an abnormal point, let the students learn the real data instead of the teacher's "half right and half wrong" data. Here, $v$ is larger than 1, indicating that when the student is not good enough, the loss will be larger. In addition, $\pmb{m}$ is the data except the outlier, $\mu$ is the parameter judging whether the point is noise or not, and $b$ is the gap between the performance of teacher and the student that is determined by the researcher. In addition, $||\bullet||^2_2$ is the square of the euclidean norm. It follows that the ARD policy can be summarized in Algorithm \ref{ard}.

\begin{algorithm}
        \caption{ARD}
        \begin{algorithmic}[1]
            \Require Traing Data $\mathbb{\pmb{D}}$
            \Ensure $L_{distill}$

            \For {$\pmb{d} \in \mathbb{\pmb{D}}$}
                \If {$||\pmb{d}-\pmb{O}_{teacher}||_2 > \mu$}
                    \State $L_{distill} = ||\pmb{O}_{student}-\pmb{O}_{teacher}||^2_2 $
                \Else
                    \State $L_{distill} = f$
                    \State \# $\pmb{m}$ is the $\mathbb{\pmb{D}}$ minus the outliers

                    \If {$||\pmb{O}_{student}-\pmb{m}||^2_2 + b > ||\pmb{O}_{teacher}-\pmb{m}||^2_2$}
                        \State $f = v ||\pmb{O}_{student}, \pmb{m}||^2_2 $
                    \Else
                        \State $f = ||\pmb{O}_{student}, \pmb{m}||^2_2$
                    \EndIf

                \EndIf

            \EndFor
        \end{algorithmic}
        \label{ard}
    \end{algorithm}

\section{Experimental Results}
\label{sec:Experimental Results}

We implemented our system on a PC with an Intel Core i7 (3.5GHz) CPU and a FACEGOOD P1 helmet camera (recording 1920 $\times$ 1080 images at 60 fps).

Fig. \ref{result} in the first line show six kinds of expression of a girl. They are anger, closing her mouth, pout, closing her right eye, puckering up, closing her left eye. The images in the second line are corresponding images with 2d landmarks marked.
It can be seen that these landmarks are accurately marked on human's lip, eyes, nose, eyebrow with almost no deviation. And the tracking is real time, with nearly no time lag, i.e landmarks adjust their coordinates according to changes in facial expression.
The third line shows six kinds of expression of the driven digital man. Compared with the first line, it is found that these figures completely restore the same expression of people, and the expression is also very delicate and lifelike.
In particular, eye and lip movements have exactly the same amplitude. For example, the range of eye movements when you close your eyes, and the range of lip movements when you pout.

\begin{figure}
\centering
  \includegraphics[scale=0.65]{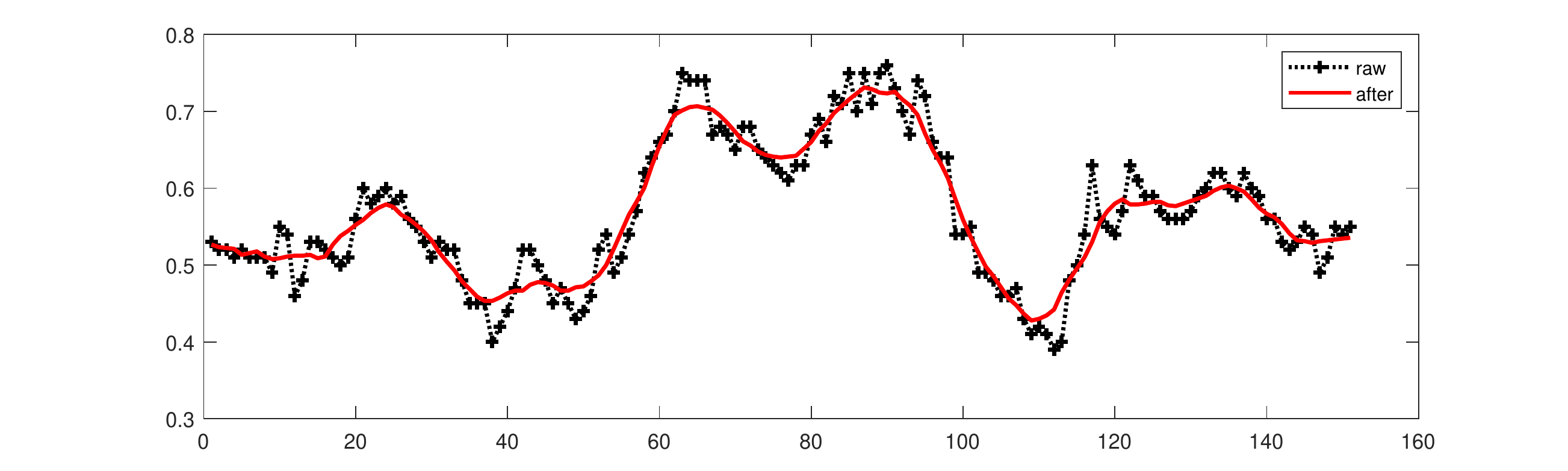}\\
  \caption{The curve comparison of whether using a hybrid filter on blendshape weight.}\label{bs weight}
\end{figure}

\begin{figure}
\centering
  \includegraphics[scale=0.65]{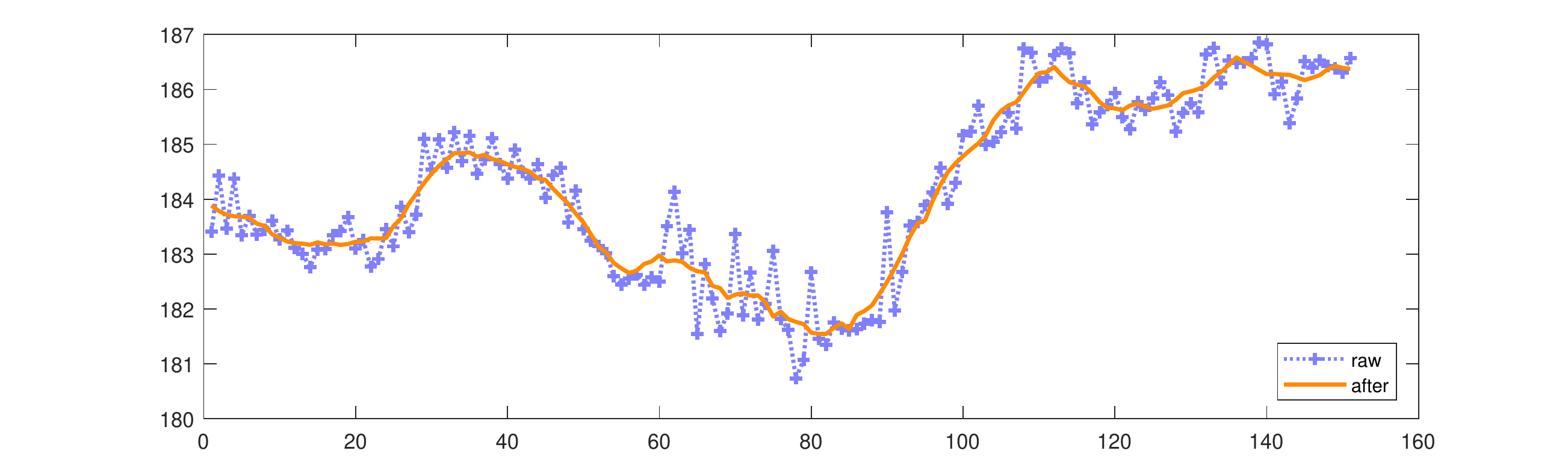}\\
  \caption{The curve comparison of whether using a hybrid filter on landmarks.}\label{landmark}
\end{figure}

Fig. \ref{bs weight} and Fig. \ref{landmark} compare whether blendshape weights and landmarks have passed through our hybrid filter. As can be seen, the filtered curve can effectively reduce jitter. In addition, the filtered curves have almost no time lag, so they can achieve the real-time level. And they also maintain the peak height, so they will not produce amplitude loss.

Table. \ref{table_time} shows the time it takes to process an image on three different CPUs, showing that our pipeline can process images at 70 FPS.

\begin{table}[!t]  \caption{Difference Equipment Running Time}
\centering

 \label{table_time}

 \begin{tabular}{cc}

\toprule

  CPU & Cost Time(ms)\\

\midrule

  i5-10400F & 20   \\

  i7-11700 &  13  \\

  i9-9900K & 13   \\

  \bottomrule

\end{tabular}

\end{table}

\section{Conclusion}
\label{sec:Conclusion}

We use the FACEGOOD P1 head-mounted infrared camera to avoid the effects of light and severe shaking caused by body movement. On this basis, an end-to-end blendshape weighting network based on neural network was developed, which can achieve 70FPS real time computing on the CPU. This method can retarget the actor's expression to any 3d models.

\bibliographystyle{unsrtnat}
\bibliography{references}  

\begin{thebibliography}{36}
\providecommand{\natexlab}[1]{#1}
\providecommand{\url}[1]{\texttt{#1}}
\expandafter\ifx\csname urlstyle\endcsname\relax
  \providecommand{\doi}[1]{doi: #1}\else
  \providecommand{\doi}{doi: \begingroup \urlstyle{rm}\Url}\fi

\bibitem[Essa et~al.(1996)Essa, Basu, Darrell, and Pentland]{essa1996modeling}
Irfan Essa, Sumit Basu, Trevor Darrell, and Alex Pentland.
\newblock Modeling, tracking and interactive animation of faces and
  heads//using input from video.
\newblock In \emph{Proceedings Computer Animation'96}, pages 68--79. IEEE,
  1996.

\bibitem[Li et~al.(2018{\natexlab{a}})Li, Zeng, Shan, and Chen]{li2018patch}
Yong Li, Jiabei Zeng, Shiguang Shan, and Xilin Chen.
\newblock Patch-gated cnn for occlusion-aware facial expression recognition.
\newblock In \emph{2018 24th International Conference on Pattern Recognition
  (ICPR)}, pages 2209--2214. IEEE, 2018{\natexlab{a}}.

\bibitem[Li et~al.(2018{\natexlab{b}})Li, Zeng, Shan, and
  Chen]{li2018occlusion}
Yong Li, Jiabei Zeng, Shiguang Shan, and Xilin Chen.
\newblock Occlusion aware facial expression recognition using cnn with
  attention mechanism.
\newblock \emph{IEEE Transactions on Image Processing}, 28\penalty0
  (5):\penalty0 2439--2450, 2018{\natexlab{b}}.

\bibitem[Lewis et~al.(2014)Lewis, Anjyo, Rhee, Zhang, Pighin, and
  Deng]{lewis2014practice}
John~P Lewis, Ken Anjyo, Taehyun Rhee, Mengjie Zhang, Frederic~H Pighin, and
  Zhigang Deng.
\newblock Practice and theory of blendshape facial models.
\newblock \emph{Eurographics (State of the Art Reports)}, 1\penalty0
  (8):\penalty0 2, 2014.

\bibitem[Mpiperis et~al.(2008)Mpiperis, Malassiotis, and
  Strintzis]{mpiperis2008bilinear}
Iordanis Mpiperis, Sotiris Malassiotis, and Michael~G Strintzis.
\newblock Bilinear models for 3-d face and facial expression recognition.
\newblock \emph{IEEE Transactions on Information Forensics and Security},
  3\penalty0 (3):\penalty0 498--511, 2008.

\bibitem[Xiao et~al.(2004)Xiao, Baker, Matthews, Kanade, et~al.]{xiao2004real}
Jing Xiao, Simon Baker, Iain Matthews, Takeo Kanade, et~al.
\newblock Real-time combined 2d+ 3d active appearance models.
\newblock In \emph{CVPR (2)}, pages 535--542, 2004.

\bibitem[Weise et~al.(2009)Weise, Li, Van~Gool, and Pauly]{weise2009face}
Thibaut Weise, Hao Li, Luc Van~Gool, and Mark Pauly.
\newblock Face/off: Live facial puppetry.
\newblock In \emph{Proceedings of the 2009 ACM SIGGRAPH/Eurographics Symposium
  on Computer animation}, pages 7--16, 2009.

\bibitem[Li et~al.(2013)Li, Yu, Ye, and Bregler]{li2013realtime}
Hao Li, Jihun Yu, Yuting Ye, and Chris Bregler.
\newblock Realtime facial animation with on-the-fly correctives.
\newblock \emph{ACM Trans. Graph.}, 32\penalty0 (4):\penalty0 42--1, 2013.

\bibitem[Laine et~al.(2017)Laine, Karras, Aila, Herva, Saito, Yu, Li, and
  Lehtinen]{laine2017production}
Samuli Laine, Tero Karras, Timo Aila, Antti Herva, Shunsuke Saito, Ronald Yu,
  Hao Li, and Jaakko Lehtinen.
\newblock Production-level facial performance capture using deep convolutional
  neural networks.
\newblock In \emph{Proceedings of the ACM SIGGRAPH/Eurographics symposium on
  computer animation}, pages 1--10, 2017.

\bibitem[Weise et~al.(2011)Weise, Bouaziz, Li, and Pauly]{weise2011realtime}
Thibaut Weise, Sofien Bouaziz, Hao Li, and Mark Pauly.
\newblock Realtime performance-based facial animation.
\newblock \emph{ACM transactions on graphics (TOG)}, 30\penalty0 (4):\penalty0
  1--10, 2011.

\bibitem[Klaudiny et~al.(2017)Klaudiny, McDonagh, Bradley, Beeler, and
  Mitchell]{klaudiny2017real}
Martin Klaudiny, Steven McDonagh, Derek Bradley, Thabo Beeler, and Kenny
  Mitchell.
\newblock Real-time multi-view facial capture with synthetic training.
\newblock In \emph{Computer Graphics Forum}, volume~36, pages 325--336. Wiley
  Online Library, 2017.

\bibitem[Bradley et~al.(2010)Bradley, Heidrich, Popa, and
  Sheffer]{bradley2010high}
Derek Bradley, Wolfgang Heidrich, Tiberiu Popa, and Alla Sheffer.
\newblock High resolution passive facial performance capture.
\newblock In \emph{ACM SIGGRAPH 2010 papers}, pages 1--10. 2010.

\bibitem[Cao et~al.(2005)Cao, Tien, Faloutsos, and Pighin]{cao2005expressive}
Yong Cao, Wen~C Tien, Petros Faloutsos, and Fr{\'e}d{\'e}ric Pighin.
\newblock Expressive speech-driven facial animation.
\newblock \emph{ACM Transactions on Graphics (TOG)}, 24\penalty0 (4):\penalty0
  1283--1302, 2005.

\bibitem[Karras et~al.(2017)Karras, Aila, Laine, Herva, and
  Lehtinen]{karras2017audio}
Tero Karras, Timo Aila, Samuli Laine, Antti Herva, and Jaakko Lehtinen.
\newblock Audio-driven facial animation by joint end-to-end learning of pose
  and emotion.
\newblock \emph{ACM Transactions on Graphics (TOG)}, 36\penalty0 (4):\penalty0
  1--12, 2017.

\bibitem[Ren et~al.(2019)Ren, Ruan, Tan, Qin, Zhao, Zhao, and
  Liu]{ren2019fastspeech}
Yi~Ren, Yangjun Ruan, Xu~Tan, Tao Qin, Sheng Zhao, Zhou Zhao, and Tie-Yan Liu.
\newblock Fastspeech: Fast, robust and controllable text to speech.
\newblock \emph{arXiv preprint arXiv:1905.09263}, 2019.

\bibitem[Liu et~al.(2008)Liu, Weissenfeld, Ostermann, and Luo]{liu2008robust}
Kang Liu, Axel Weissenfeld, Joern Ostermann, and Xinghan Luo.
\newblock Robust aam building for morphing in an image-based facial animation
  system.
\newblock In \emph{2008 IEEE International Conference on Multimedia and Expo},
  pages 933--936. IEEE, 2008.

\bibitem[Liu and Ostermann(2009)]{liu2009optimization}
Kang Liu and Joern Ostermann.
\newblock Optimization of an image-based talking head system.
\newblock \emph{EURASIP journal on audio, speech, and music processing},
  2009:\penalty0 1--13, 2009.

\bibitem[Liu and Ostermann(2011)]{liu2011realistic}
Kang Liu and Joern Ostermann.
\newblock Realistic facial expression synthesis for an image-based talking
  head.
\newblock In \emph{2011 IEEE International Conference on Multimedia and Expo},
  pages 1--6. IEEE, 2011.

\bibitem[Taylor et~al.(2016)Taylor, Kato, Milner, and
  Matthews]{taylor2016audio}
Sarah Taylor, Akihiro Kato, Ben Milner, and Iain Matthews.
\newblock Audio-to-visual speech conversion using deep neural networks.
\newblock 2016.

\bibitem[Chen et~al.(2013)Chen, Wu, Shi, Tong, and Chai]{chen2013accurate}
Yen-Lin Chen, Hsiang-Tao Wu, Fuhao Shi, Xin Tong, and Jinxiang Chai.
\newblock Accurate and robust 3d facial capture using a single rgbd camera.
\newblock In \emph{Proceedings of the IEEE International Conference on Computer
  Vision}, pages 3615--3622, 2013.

\bibitem[Cao et~al.(2013{\natexlab{a}})Cao, Weng, Lin, and Zhou]{cao20133d}
Chen Cao, Yanlin Weng, Stephen Lin, and Kun Zhou.
\newblock 3d shape regression for real-time facial animation.
\newblock \emph{ACM Transactions on Graphics (TOG)}, 32\penalty0 (4):\penalty0
  1--10, 2013{\natexlab{a}}.

\bibitem[Cao et~al.(2013{\natexlab{b}})Cao, Weng, Zhou, Tong, and
  Zhou]{cao2013facewarehouse}
Chen Cao, Yanlin Weng, Shun Zhou, Yiying Tong, and Kun Zhou.
\newblock Facewarehouse: A 3d facial expression database for visual computing.
\newblock \emph{IEEE Transactions on Visualization and Computer Graphics},
  20\penalty0 (3):\penalty0 413--425, 2013{\natexlab{b}}.

\bibitem[Cao et~al.(2014)Cao, Hou, and Zhou]{cao2014displaced}
Chen Cao, Qiming Hou, and Kun Zhou.
\newblock Displaced dynamic expression regression for real-time facial tracking
  and animation.
\newblock \emph{ACM Transactions on graphics (TOG)}, 33\penalty0 (4):\penalty0
  1--10, 2014.

\bibitem[Chuang and Bregler(2002)]{chuang2002performance}
Erika Chuang and Chris Bregler.
\newblock Performance driven facial animation using blendshape interpolation.
\newblock \emph{Computer Science Technical Report, Stanford University},
  2\penalty0 (2):\penalty0 3, 2002.

\bibitem[Khanam and Mufti(2007)]{khanam2007intelligent}
Assia Khanam and Muid Mufti.
\newblock Intelligent expression blending for performance driven facial
  animation.
\newblock \emph{IEEE Transactions on Consumer Electronics}, 53\penalty0
  (2):\penalty0 578--584, 2007.

\bibitem[Deena and Galata(2009)]{deena2009speech}
Salil Deena and Aphrodite Galata.
\newblock Speech-driven facial animation using a shared gaussian process latent
  variable model.
\newblock In \emph{International Symposium on Visual Computing}, pages 89--100.
  Springer, 2009.

\bibitem[Meng and Wen(2019)]{meng2019lstm}
Hsien-Yu Meng and Jiangtao Wen.
\newblock Lstm-based facial performance capture using embedding between
  expressions.
\newblock In \emph{Proceedings of SAI Intelligent Systems Conference}, pages
  211--226. Springer, 2019.

\bibitem[Habermann et~al.(2019)Habermann, Xu, Zollhoefer, Pons-Moll, and
  Theobalt]{habermann2019livecap}
Marc Habermann, Weipeng Xu, Michael Zollhoefer, Gerard Pons-Moll, and Christian
  Theobalt.
\newblock Livecap: Real-time human performance capture from monocular video.
\newblock \emph{ACM Transactions On Graphics (TOG)}, 38\penalty0 (2):\penalty0
  1--17, 2019.

\bibitem[Kettern et~al.(2015)Kettern, Hilsmann, and
  Eisert]{kettern2015temporally}
Markus Kettern, Anna Hilsmann, and Peter Eisert.
\newblock Temporally consistent wide baseline facial performance capture via
  image warping.
\newblock In \emph{VMV}, pages 95--102, 2015.

\bibitem[Zhang et~al.(2014)Zhang, Tjondronegoro, and Chandran]{zhang2014random}
Ligang Zhang, Dian Tjondronegoro, and Vinod Chandran.
\newblock Random gabor based templates for facial expression recognition in
  images with facial occlusion.
\newblock \emph{Neurocomputing}, 145:\penalty0 451--464, 2014.

\bibitem[Pan et~al.(2019)Pan, Wang, and Xia]{pan2019occluded}
Bowen Pan, Shangfei Wang, and Bin Xia.
\newblock Occluded facial expression recognition enhanced through privileged
  information.
\newblock In \emph{Proceedings of the 27th ACM International Conference on
  Multimedia}, pages 566--573, 2019.

\bibitem[Apple()]{arkit}
Apple.
\newblock Arkit.
\newblock
  \url{https://developer.apple.com/documentation/arkit/arfaceanchor/blendshapelocation/2928266-mouthclose}.

\bibitem[Szegedy et~al.(2017)Szegedy, Ioffe, Vanhoucke, and
  Alemi]{szegedy2017inception}
Christian Szegedy, Sergey Ioffe, Vincent Vanhoucke, and Alexander~A Alemi.
\newblock Inception-v4, inception-resnet and the impact of residual connections
  on learning.
\newblock In \emph{Thirty-first AAAI conference on artificial intelligence},
  2017.

\bibitem[Kalman(1960)]{kalman1960new}
Rudolph~Emil Kalman.
\newblock A new approach to linear filtering and prediction problems.
\newblock 1960.

\bibitem[Press and Teukolsky(1990)]{press1990savitzky}
William~H Press and Saul~A Teukolsky.
\newblock Savitzky-golay smoothing filters.
\newblock \emph{Computers in Physics}, 4\penalty0 (6):\penalty0 669--672, 1990.

\bibitem[Kingma and Ba(2014)]{kingma2014adam}
Diederik~P Kingma and Jimmy Ba.
\newblock Adam: A method for stochastic optimization.
\newblock \emph{arXiv preprint arXiv:1412.6980}, 2014.

\end{thebibliography}






\end{document}